\pdfoutput=1

\documentclass[11pt]{article}

\usepackage[final]{acl}

\usepackage{times}
\usepackage{latexsym}

\usepackage[T1]{fontenc}

\usepackage[utf8]{inputenc}

\usepackage{microtype}

\usepackage{inconsolata}
\usepackage{subfigure}
\usepackage{amsmath}
\usepackage{amssymb}
\usepackage{mathtools}
\usepackage{amsthm}
\usepackage{booktabs} 
\usepackage{graphicx}
\usepackage{hyperref}
\usepackage{url}
\usepackage{pgfplots}
\usetikzlibrary{pgfplots.groupplots}
\usepackage{wrapfig, lipsum, booktabs}
\pgfplotsset{compat=1.3}
\usepackage{tikz}
\usepackage{listings}
\lstdefinestyle{prompt}{
  basicstyle=\ttfamily\scriptsize,
  breaklines=false,
  frame=single,
  columns=fullflexible,
  escapeinside={(*}{*)},
  backgroundcolor=\color{gray!5!white},
}

\usetikzlibrary{patterns}
\usepackage{pifont} 
\usepackage[T1]{fontenc}
\newcommand{\okmark}{{\textbf{\textcolor[rgb]{0.1, 0.5, 0.1}{$\checkmark$}}}}
\newcommand{\ngmark}{{\textbf{\color{red}{\ding{55}}}}}

\definecolor{battleshipgrey}{rgb}{0.3, 0.3, 0.3}
\definecolor{brilliantrose}{rgb}{1.0, 0.33, 0.64}
\definecolor{americanrose}{rgb}{1.0, 0.01, 0.24}
\definecolor{jweigreen}{rgb}{0,0.45,0.24}
\definecolor{bluegray}{rgb}{0.1, 0.1, 0.4}
\definecolor{ao(english)}{rgb}{0.0, 0.5, 0.0}
\definecolor{blanchedalmond}{rgb}{1.0, 0.92, 0.8}
\definecolor{atomictangerine}{rgb}{1.0, 0.6, 0.4}
\definecolor{chocolate(web)}{rgb}{0.82, 0.41, 0.12}
\definecolor{bananayellow}{rgb}{1.0, 0.88, 0.21}
\definecolor{goldenbrown}{rgb}{0.6, 0.4, 0.08}
\definecolor{aliceblue}{rgb}{0.94, 0.97, 1.0}
\definecolor{beige}{rgb}{0.96, 0.96, 0.86}
\definecolor{babyblue}{rgb}{0.54, 0.81, 0.94}
\definecolor{camel}{rgb}{0.76, 0.6, 0.42}
\definecolor{cinnamon}{rgb}{0.82, 0.41, 0.12}

\usepackage{pgfplots}
\usetikzlibrary{pgfplots.groupplots}
\pgfplotsset{compat=1.3}
\usepackage{pifont}
\usepackage{microtype}

\title{You Only Look at Screens: Multimodal Chain-of-Action Agents}

\author{Zhuosheng Zhang\textsuperscript{1}\thanks{Work done at Amazon Web Services.}, Aston Zhang\textsuperscript{2}$^{*}$\\
\textsuperscript{1} School of Electronic Information and Electrical Engineering,\\ Shanghai Jiao Tong University\\
\textsuperscript{2} GenAI, Meta \\
\texttt{zhangzs@sjtu.edu.cn, az@astonzhang.com}\\
}

\begin{document}
\maketitle
\begin{abstract}
Autonomous graphical user interface (GUI) agents aim to facilitate task automation by interacting with the user interface without manual intervention. Recent studies have investigated eliciting the capabilities of large language models (LLMs) for effective engagement in diverse environments. To align with the input-output requirement of LLMs, most existing approaches are developed under a sandbox setting where they rely on external tools and application-specific APIs to parse the environment into textual elements and interpret the predicted actions. Consequently, those approaches often grapple with inference inefficiency and error propagation risks. To mitigate the challenges, we introduce Auto-GUI, a multimodal solution that directly interacts with the interface, bypassing the need for environment parsing or reliance on application-dependent APIs. Moreover, we propose a chain-of-action technique---leveraging a series of intermediate previous action histories and future action plans---to help the agent decide what action to execute. We evaluate our approach on a new device-control benchmark AITW with 30$K$ unique instructions, spanning multi-step tasks such as application operation, web searching, and web shopping. Experimental results show that Auto-GUI achieves state-of-the-art performance with an action type prediction accuracy of 90\% and an overall action success rate of 74\%. Code is publicly available at \texttt{https://github.com/cooelf/Auto-GUI}.

\end{abstract}

\section{Introduction}


\begin{figure*}[t]
  \begin{center}
   \includegraphics[width=1.0\textwidth]{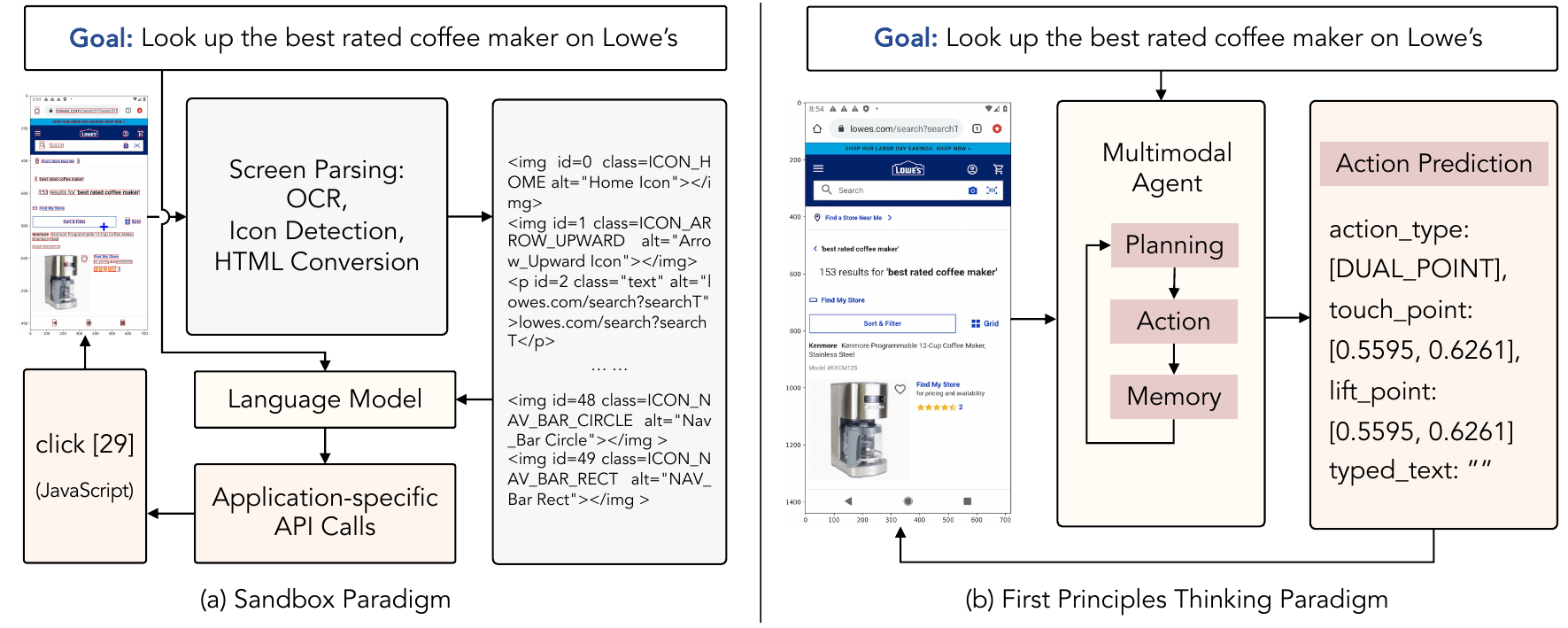}
  \end{center}
  \vspace{-3mm}
  \caption{Comparison of GUI agent paradigms. {The sandbox paradigm depends on the intermediate transformation between environments and agents, i.e., needing access to intermediate environment parsing or interval application-dependent APIs. In contrast, our first principles thinking paradigm allows direct interactions on the screen without intermediate transformation. Details of the action types and action points are presented in Section \ref{sec:norm}.}}
  \label{fig:paradigm}
  \vspace{-3mm}
\end{figure*}

Building intelligent autonomous agents that are capable of task planning, decision making, and action execution in a particular environment is a long-standing goal of artificial intelligence (AI) \citep{searle1969speech,wooldridge1995intelligent,maes1995agents,hendler1999there}. 
The advent of large language models (LLMs) \citep{brown2020language,chowdhery2022palm,openai2023gpt4} has flourished promising opportunities for developing autonomous agents to assist users in completing tasks in distinct environments such as operation systems, specific applications, and web browsers \citep{act-1,rawles2023android,liu2023agentbench,zhou2023webarena,wang2023survey,koh2024visualwebarena,gao2023assistgui,yang2023appagent}.

Recent studies have explored prompt engineering \citep{autogpt,babyagi,agentgpt,sumers2023cognitive,liu2023agentbench} and fine-tuning techniques \citep{rawles2023android,wen2023empowering,sun2022meta} to elicit the capability of language models to execute actions in interactive environments. However, 
there are at least two major challenges that have limited real-world applications of autonomous agents.

First, existing approaches commonly rely on external tools such as optical character recognition (OCR) and icon detectors \citep{zhang2021screen,sunkara2022towards} to parse the environment into textual elements (e.g., HTML layouts) as inputs to a language model (Figure \ref{fig:paradigm}(a)) \citep{rawles2023android}. On the one hand, the parsed elements generate lengthy inputs, thus leading to inference {inefficiency}. Since computational latency is a key measure in deployment, using lengthy inputs would increase inference cost and may even exceed the input length limit of the language model. On the other hand, parsing the visual environment into textual elements may also be prone to error propagation or information loss because parsing mistakes are inevitable using external tools.

Second, most existing approaches are under the sand-box setting that requires accessing internal APIs to interact with the environment \citep{zhou2023webarena,gur2023real}, e.g., using a JavaScript element selection on a webpage or a Python interpreter to execute actions. However in practice, the API interface is often inaccessible in third-party applications (Apps).

These challenges have motivated more advanced techniques that are capable of \emph{first principles thinking} \citep{aristotle,irwin1989aristotle}---allowing direct interactions on the screen without needing access to intermediate environment parsing or interval application-dependent APIs (Figure \ref{fig:paradigm}(b)). To address the challenges, we introduce \textbf{Auto-GUI}, a multimodal approach that directly interacts with the graphical user interfaces (GUIs). 
To further strengthen the agent's action prediction capability,
we propose a novel \textbf{chain-of-action} technique,
where a chain of action is a series of intermediate previous action histories and future action plans.

We evaluate Auto-GUI on a new device-control benchmark AITW \citep{rawles2023android} with 30$K$ unique instructions, spanning multi-step tasks of application operation, web searching, and web shopping. Experimental results show that Auto-GUI achieves state-of-the-art performance with an action type prediction accuracy of 90\% and an action success rate of 74\%.

In summary, our work makes the following technical contributions:

(i) We introduce Auto-GUI, a multimodal agent for autonomous GUI control that can directly interact with the screens, thus circumventing the constraints of environment parsing and application-specific API access. 

(ii) We propose a chain-of-action technique that leverages the previously executed actions and future action plans to help the agent decide what action to execute at each step.

(iii) Auto-GUI achieves state-of-the-art performance with an action type prediction accuracy of 90\% and an action success rate of 74\%. Notably, Auto-GUI is able to infer an action in less than one second.

 \begin{figure*}[t]
  \begin{center}
 \includegraphics[width=1.0\textwidth]{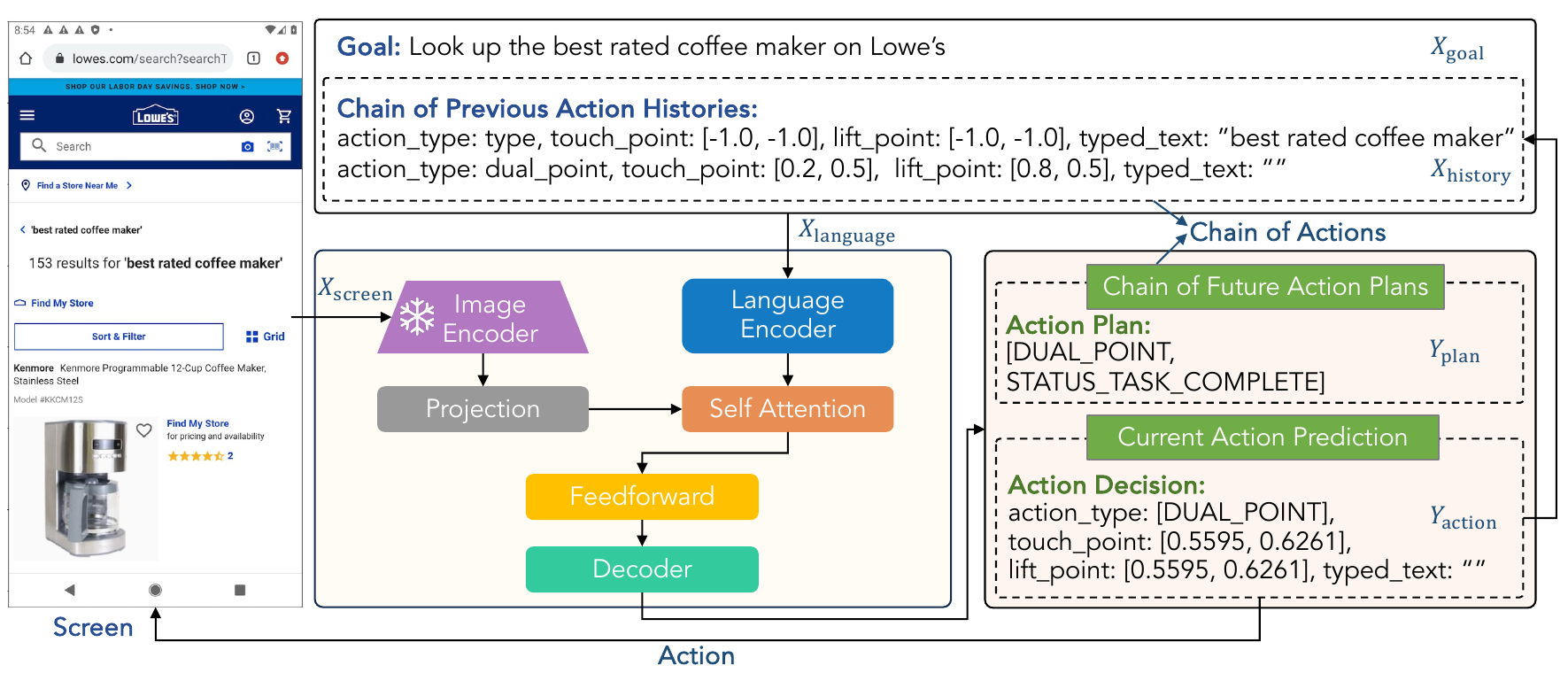}
  \end{center}
  \vspace{-3mm}
  \caption{Model architecture of Auto-GUI. A chain of action consists of a chain of previous action histories $X_{\textrm{history}}$ and a chain of future action plans $Y_{\textrm{plan}}$ in the illustration. }
   \vspace{-3mm}
  \label{fig:approach}
\end{figure*}

\section{Related Work}
Our work falls into the field of language agents. This section will first review the recent progress in building language agents and then discuss the approaches to conduct user interface control with language agents.

\subsection{Language Agents}
Language agents refer to those agents that can follow user instructions and interact with environments to complete tasks. Such agents expand the landscape of language models to compete in specific fields, including application operation, web searching, and web shopping. There are two popular types of language agents, autonomous agents and communicative agents. Autonomous agents aim to assist humans to achieve specific goals in the real world. Typical examples of autonomous agents are AutoGPT \citep{autogpt}, BabyAGI \citep{babyagi}, and AgentGPT \citep{agentgpt}. In contrast, communicative agents are personalized and socialized agents \citep{park2023generative,wang2023voyager,zhu2023ghost,hong2023metagpt} with human behaviors that can communicate and collaborate with each other. They are often deployed in immersive environments. 

Inspired by the potential in real-world applications, this work focuses on autonomous agents, especially those working in mobile devices. We aim to assist users by completing multi-step tasks (e.g., manipulating Apps, web shopping, and question answering) without any manual intervention. Given a user instruction in natural language, the agent is required to interpret the instruction and execute actions by directly controlling its user interface. Due to the requirement in real-world applications, the agent is expected to be both effective and efficient.

\subsection{GUI Control with Natural Language}
Recently, LLMs have shown promise in building autonomous GUI agents with abilities of instruction following \citep{sanh2021multitask,taori2023stanford,chiang2023vicuna} and chain-of-thought (CoT) prompting \citep{nye2022show,cot_wei}. Especially, CoT prompting \citep{cot_wei,kojima2022large,zhang2022automatic} elicit LLMs' capacities of step-by-step planning, decision making, and action execution. Those capacities have been shown to be effective in GUI control tasks \citep{rawles2023android}.

However, the task environments are GUIs instead of natural language that LLMs can process directly. Therefore, the GUI states and actions are required to be converted to textual formats to be applicable to LLMs. For example, it is feasible to parse the GUI screens by icon recognition and OCR \citep{zhang2021screen,sunkara2022towards,song2023navigating} and organize the parsed elements into HTML layouts. As a compromise, existing approaches are restricted in a sandbox setting where they rely on external tools \citep{rawles2023android,wen2023empowering} and application-specific APIs \citep{zhou2023webarena,gur2023real} for environment parsing and action interpretation; thus, commonly suffer from inference {inefficiency} and error propagation. Although there are studies that have considered multimodal architecture for processing inputs in different modalities \citep{sun2022meta,yan2023gpt}, those studies still rely on fine-grained environment parsing to ensure competitive performance. 
As GUI tasks have shown prerequisites to fine-grained grounding to the GUI contents, more recent concurrent studies \citep{cheng2024seeclick,hong2023cogagent} have explored GUI grounding
pre-training to improve the agent's performance.

In contrast to the studies above, this work is established upon first principles thinking, which directly reads GUI without additional environment parsing and provides the action (e.g., action type, gesture coordinate, and typed text) that can be efficiently executed without needing any extra APIs. Concretely, our approach diverges from existing studies in several key aspects.

(i) Conceptually, Auto-GUI represents a novel approach as a multimodal agent designed for autonomous GUI control, enabling direct interaction with screens. This approach effectively bypasses the need for environment parsing and access to application-specific APIs, thus enhancing adaptability and versatility.

(ii) Methodologically, our proposal introduces a chain-of-action technique. This method capitalizes on previously executed actions and future action plans to inform the decision-making process of the agent at each step.

(iii) As a scientific contribution, our paper articulates that a unified multimodal model out of first principles thinking can serve as a strong autonomous agent evidenced by state-of-the-art performance. In addition, we facilitate a series of studies to understand the contributing factors regarding the vision features and model scale.

\section{Methodology}
 In this section, we will first introduce the basic concepts for the GUI control task and then describe the design of our proposed Auto-GUI framework.

\subsection{Problem Formalization}
Given a user instruction (also known as a \textit{goal}), the agent needs to complete the task with multiple steps of interactions. The entire process is called an \textit{episode}, which is composed of a series of \textit{screens}. For each step in the episode, the agent will be provided with a screenshot, and the agent is required to predict the action until the task is complete. Detailed examples can be found in Appendix \ref{app:example}.

\subsection{Framework Overview}
Auto-GUI is a multimodal agent that decides what action to take given the input screenshot and a user instruction. To empower the agent's decision making capability, we introduce a chain-of-action approach by leveraging a series of intermediate previous action histories and future action plans to predict actions.

The model architecture of Auto-GUI is illustrated in Figure \ref{fig:approach}. On a high level, Auto-GUI consists of three stages. 
First, we acquire encoded features from both vision and language inputs.  Specifically, the vision input, i.e., a screenshot, is encoded by a frozen vision encoder. Meanwhile, the language input, consisting of the goal and a chain of previous action histories---each history contains a tuple \{action type, touch point, lift point, and typed text\}, is encoded by a language encoder. Second, the encoded vision and language representations are integrated by a self-attention module. Third, the fused representation is fed to the decoder to generate a chain of future action plans (i.e., action types to execute in future steps) followed by action prediction.
A chain of action consists of two parts in the procedure above: a chain of previous action histories on the input side and a chain of future action plans on the output side. In the following, we describe the entire procedure in detail.

\paragraph{Encoding} 
Suppose that an episode consists of $k$ steps of interactions. Given a screenshot $X_{\textrm{screen}} \in \mathbb{R}^{h \times w \times 3}$ with height $h$ and width $w$ at step $t \in [1, k]$, we first feed it to a frozen image
encoder (e.g., BLIP-2 \citep{li2023blip}) and extract vision features $H_{\textrm{screen}} \in \mathbb{R}^{1 \times d_{s}}$ where $d_{s}$ is the dimension of the vision features. Additionally, we leverage a language encoder to extract the language features $H_{\textrm{language}} \in \mathbb{R}^{n \times d_{l}}$ of the input goal $X_{\textrm{goal}}$ where $n$ is the number of tokens and $d_{l}$ is the dimension of the language features. If $t > 1$, there will be a chain-of-action history already executed before step $t$. We denote the chain of action histories as $X_{\textrm{history}} = [m_1, \dots, m_t]$ where $m_i$ contains a tuple of action type, touch point, lift point, and typed text.\footnote{Following the design philosophy of AITW \citep{rawles2023android}, the agent is presumed to execute a sequence of actions until the task is completed. For practical use, it is feasible to halt the agent using straightforward rules when required.} Otherwise, if $t=1$, $X_{\textrm{history}}$ will be set empty:
\begin{equation}
X_{\textrm{history}} = \begin{cases}
 [m_1, \dots, m_t], & \text{ if } t > 1\\
 \text{<empty>}, & \text{otherwise}
\end{cases}
\end{equation}%

We concatenate $X_{\textrm{goal}}$ and $X_{\textrm{history}}$ as the input to the language encoder: 
$
X_{\textrm{language}} = \{X_{\textrm{goal}}, X_{\textrm{history}}\}.
$

Then, we obtain the encoded representations of the vision and language inputs as follows:
\begin{eqnarray}
    H_{\textrm{screen}} & = & \textrm{VisionExtractor}(X_{\textrm{screen}}),\\
    H^{'}_{\textrm{screen}}  & = & W H_{\textrm{screen}},\\
    H_{\textrm{language}} & = & \textrm{LanguageEncoder}(X_{\textrm{language}}),
	\label{eq:extractor}
\end{eqnarray}
where $W$ is a trainable projection matrix to convert $H_{\textrm{screen}}$ into the same dimensionality as $H_{\textrm{language}}$.

\paragraph{Interaction} 
We correlate $H^{'}_{\textrm{screen}}$ and $H_{\textrm{language}}$ with a single-head self-attention network \citep{vaswani2017attention},\footnote{It is possible to use a unified multimodal Transformer architecture \citep{fuyu-8b}. Here we opt for a single-layer interaction to maintain the lightweight nature of our method, with parameters less than 1 billion. This design choice also ensures flexibility to accommodate new modalities in the future, such as integrating additional feature extractors.} where the query (${Q}$), key (${K}$), and value (${V}$) are $H_{\textrm{language}}$, $H^{'}_{\textrm{screen}}$, and $H^{'}_{\textrm{screen}}$, respectively. The attention output $H_{\textrm{screen}}^\textrm{attn} \in \mathbb{R}^{n \times d}$ is defined as:
$H_{\textrm{screen}}^\textrm{attn} = \textrm{Softmax}(\frac{{Q}{{K}^{\top}}}{\sqrt{d_k}}){V}$, where $d_k$ is the same as the dimension of $H_{\textrm{language}}$ because a single head is used. 

Then, a gated fusion \citep{zhang2020neural,wu2021good,zhang2023multimodal} is adopted to fuse $H_{\textrm{language}}$ and $H_{\textrm{screen}}^\textrm{attn}$. We have the fused output $H_{\textrm{fuse}} \in \mathbb{R}^{n \times d}$ by:
\begin{eqnarray}
	&\lambda  = \textrm{Sigmoid}(W_{l}{H_{\textrm{language}}} + W_{v}{H_{\textrm{vision}}^\textrm{attn}}), \label{eq:gate}\\
	&H_{\textrm{fuse}} = (1 - \lambda) \cdot H_{\textrm{language}} + \lambda \cdot H_{\textrm{vision}}^\textrm{attn} ,\label{eq:gated_fusion}
\end{eqnarray}
\noindent where $W_{l}$ and $W_{v}$ are learnable parameters. 

\paragraph{Decoding}
The fused representation $H_{\textrm{fuse}}$ is fed to a Transformer decoder to generate the target predictions in a string format. The target predictions consist of a chain of future action plans $Y_{\textrm{plan}}$ and the current action prediction $Y_{\textrm{action}}$ separated by specific prompts: \{Action Plan: $Y_{\textrm{plan}}$, Action Decision: $Y_{\textrm{action}}\}$. Concretely, $Y_{\textrm{plan}}$ is a chain of action types to execute in future steps: 
$Y_{\textrm{plan}}$ = [action\_type$_{t}$, $\dots$, action\_type$_{k}$]. $Y_{\textrm{action}}$ contains four components: $Y_{\textrm{action}}$ = \{``action\_type'': <action\_type>, ``touch\_point'': <touch\_point>, ``lift\_point'': <lift\_point>, ``typed\_text'': <typed\_text>\}.
These four components will be explained as follows.

\subsection{Coordinate Normalization}\label{sec:norm}
Recall that a target action consists of four components: action type, touch point, lift point, and typed text. We consider six action types: \textit{dual-point gesture}, \textit{type}, \textit{go\_back}, \textit{go\_home}, \textit{enter}, and \textit{status\_complete}.  A dual-point gesture comprises a touch point and a lift point with $[y, x]$ coordinates.\footnote{We follow the $[y, x]$ coordinate format in AITW.} The gesture actions ensure a flexible action space and can represent clicks and scrolls at arbitrary locations. 
For example, a gesture action \{``touch\_point'': [0.7761, 0.7089], ``lift\_point'': [0.7761, 0.7089]\} means clicking at the coordinate [0.7761, 0.7089], while a gesture action \{``touch\_point'': [0.1898, 0.4477], ``lift\_point'': [0.8242, 0.4077]\} means scrolling down. A type action means typing a text and the text is placed in the <typed\_text> field. The other action types, i.e., go\_back, go\_home, enter, and status\_complete are system actions, whose corresponding <touch\_point>, <lift\_point> fields are filled with -1, and the <typed\_text> is empty.

We observe that high-precision coordinates are not necessary for representing a click or scroll action. Therefore, we apply normalized values of the coordinates, which helps accelerate convergence and mitigate the ambiguity of coordinates. The normalization is applied to click and scroll actions. For click actions, we keep four decimal places. For scroll actions, we first determine the scroll direction with the touch and lift points. Then, we transform the touch and lift points into fixed directional coordinates as follows: ``up'': \{[0.8, 0.5], [0.2, 0.5]\},
``down'': \{[0.2, 0.5], [0.8, 0.5]\},
``left'': \{[0.5, 0.8], [0.5, 0.2]\},
``right'': \{[0.5, 0.2], [0.5, 0.8]\}, where \{[$\cdot$], [$\cdot$]\} consists of the touch point and lift point in the first [$\cdot$] and second [$\cdot$].\footnote{The numbers are relative scales and they generalize to different screen sizes.} We provide examples of target actions in Appendix \ref{app:norm}.

\section{Experiments}
\subsection{Dataset}
We use the AITW benchmark dataset \citep{rawles2023android}. AITW is a large-scale GUI control benchmark dataset containing natural language instructions, screenshots, and actions. There are 715$K$ episodes spanning 30$K$ unique instructions, covering diverse multi-step tasks such as application operation, web searching, and web shopping, on over 350 Apps and websites. This dataset covers various device types and operation systems in varying screen resolutions to ensure generality. There are five subsets in the benchmark dataset, namely, General, Install, GoogleApps, Single, and WebShopping. 

\begin{table}[!htb]
    \centering\small
    \begin{tabular}{lllc}
    \toprule
     {Dataset}  & Episodes & Screens  &  Instructions\\
    \midrule
     General & 9,476 & 85,413 & 545 \\
     Install &  25,760 & 250,058 & 688 \\
     GoogleApps &  625,542 & 4,903,601 & 306  \\
     Single & 26,303 & 85,668 & 15,366 \\
     WebShopping & 28,061 & 365,253 & 13,473 \\
    \bottomrule
  \end{tabular}
  \caption{Dataset statistics\label{tab:data-app}.}
\end{table}

Table \ref{tab:data-app} presents the data statistics of the AITW dataset. Each subset is split episode-wise into a training, validation, and test set (80/10/10\%). More details of the subsets can be found in Appendix \ref{appendix:data}.

\subsection{Baselines}
We adopt three types of baselines, allowing for a comprehensive comparison with our approach. {The baselines encompass the in-context earning (ICL) and fine-tuning paradigms. They are based on various backbone models of different sizes.}

{(i) In-context Learning LLMs. Few-shot PaLM 2, ChatGPT (turbo-3.5) are adopted. Following prior studies \citep{rawles2023android,wang2023enabling}, we feed the LLM a textual description of the screen and a user instruction. The screen is formatted as an HTML syntax, providing the information of GUI elements derived from OCR detection and icon detection from external tools \citep{rawles2023android}. The model is required to predict an action among pre-defined actions. In addition, we report the results of the multimodal GPT-4V by taking the vision image and action history as the input based on \citet{yan2023gpt}.}

{(ii) Fine-tuned LLMs. We adopt Llama-2-7B \citep{touvron2023llama} as the baseline and fine-tune it with LoRA. We feed the model with the user instruction and the screen descriptions in HTML syntax (the same as in-context learning LLMs). The model is expected to predict the action in the same output format as in-context learning LLMs.}

{(iii) Specialized GUI Agent. We adopted the Behavioural Cloning (BC) agent, which reported the state-of-the-art performance in \citet{rawles2023android}. BC is a Transformer-based architecture that takes a task instruction, the current screen, and a stacked history of screen observations and actions as input. All the embedded representations are fused to predict the action by a decoder. There are two BC variants, BC-single and BC-history, depending on whether the model takes the screen-action history as input.}

More detailed implementation of the baselines can be found in Appendix \ref{app:baselines}.

\begin{table*}[t]
\centering
\small 
\renewcommand\tabcolsep{6pt} 
\begin{tabular}{l|cc|c|ccccc} 
\toprule
 Model & Unified & w/o Anno. & Overall & General & Install & GoogleApps & Single & WebShopping \\
 \midrule
PaLM 2-CoT & \okmark & \ngmark& 39.6	&  -  &  - &  - &  -  &\\
ChatGPT-CoT & \okmark& \ngmark &7.72 &	5.93	&4.38	&10.47	& 9.39	& 8.42\\
{GPT-4V} & \okmark& \ngmark & 52.96 & 43.01 & 46.14 & 49.18 & 78.29 & 48.18\\
\midrule
Fine-tuned Llama 2 & \ngmark & \ngmark & 28.40 & 28.56 & 35.18 & 30.99 & 27.35 & 19.92\\ 
\midrule
BC-single & \ngmark & \ngmark & 68.7	& - & - &  - &  -  &\\
BC-history & \ngmark & \ngmark& \underline{73.1} &	\underline{63.7}	& \underline{77.5}	& \underline{75.7}	& \underline{80.3}	& \underline{68.5} \\
\midrule
Auto-GUI$_\text{separate}$ & \ngmark & \okmark & 74.07 & 65.94 & \textbf{77.62} & \textbf{76.45} & 81.39 & 69.72 \\
Auto-GUI$_\text{unified}$ & \okmark & \okmark & \textbf{74.27}	& \textbf{68.24} &	{76.89} & {71.37} & \textbf{84.58} & \textbf{70.26} \\
 \bottomrule
\end{tabular}
\caption{{Main results (\%). Segment 1: in-context learning LLM baselines; Segment 2: fine-tuned Llama 2 baseline; Segment 3: specialized agent baselines; Segment 4: our Auto-GUI results. Prior published best results are marked with an \underline{underline}. ``Unified'' means a general model that can work across subsets. ``w/o Anno.'' means no screen description is needed. The PaLM-CoT and BC results are from \citet{rawles2023android}. The GPT-4V result is from \citet{yan2023gpt}. The other results are based on our own implementations. The overall score is computed as the average accuracy on all the subsets.
The best average result is in \textbf{bold} face.}}
 \label{tab:main_results}
 \vspace{-3mm}
\end{table*}

\subsection{Evaluation Measures}
We compute the screen-wise action matching score as the main evaluation measure, defined as the number of correct actions divided by the episode length (i.e., the number of screenshots). A predicted action is considered correct if the action type and dual-point gesture match the gold ones. As we described in Section \ref{sec:norm}, the gesture actions can represent the click actions and scroll actions at arbitrary locations. {Following \citet{rawles2023android}, a click action is considered correct if its touch point and lift point fall within a 14\% screen distance from the gold gestures or occur within the same detected bounding box with the gold gestures.} A scroll action is considered correct if it has the same scroll axis as the gold gesture.

The screen-wise action matching score has been shown to correlate with the task complete score estimated by human evaluations \citep{rawles2023android} and is appropriate to measure the action success rate for user instructions. Besides the overall matching score, we will also compare the click region accuracy, scroll direction accuracy, action type accuracy (e.g., clicking, scrolling, typing, etc), and typed text accuracy for a more comprehensive reference (Section \ref{sec:category}).

The evaluation criteria above apply to the BC baselines and our Auto-GUI. For the LLMs, they can only click on detected GUI elements, rather than clicking at arbitrary locations. Therefore, we consider if the clicked GUI element is matched for click actions instead of comparing dual-point gestures for LLMs.

\subsection{Implementation Details}
We adopt the encoder-decoder architecture \citep{raffel2020exploring} under $\text{small}$ (60M), $\text{base}$ (200M) and $\text{large}$ (700M) settings in our framework. We apply FLAN-Alpaca to initialize our model weights.\footnote{\url{https://github.com/declare-lab/flan-alpaca}.} The vision features are obtained by the frozen BLIP-2 encoder \citep{li2023blip} (version: blip2\_t5\_instruct). We fine-tune the models up to 10 epochs, with a learning rate of 1e-4. The maximum input sequence length is 512. The batch size is 4. Our experiments are run on 8 NVIDIA Tesla V100 32G GPUs. Training the large and base models takes 75 and 25 hours, respectively.

We develop two kinds of approaches to analyze their generalization abilities, namely Auto-GUI$_\text{separate}$, and Auto-GUI$_\text{unified}$. Specifically, Auto-GUI$_\text{separate}$ is trained and evaluated independently on each subset. Auto-GUI$_\text{unified}$ is a unified model trained on the training sets of each subset and evaluated on each test set. As the GoogleApps subset is 10-100 times larger than the other subsets, using all the training data to train a unified model would suffer from the data imbalance issue \citep{zhang2022task}. Therefore, we only use 10\% training data of GoogleApps. At the same time, the overall computation cost can also be saved by 80\%. 
We use Auto-GUI$_\text{unified}$ as the default model for analysis unless otherwise stated.

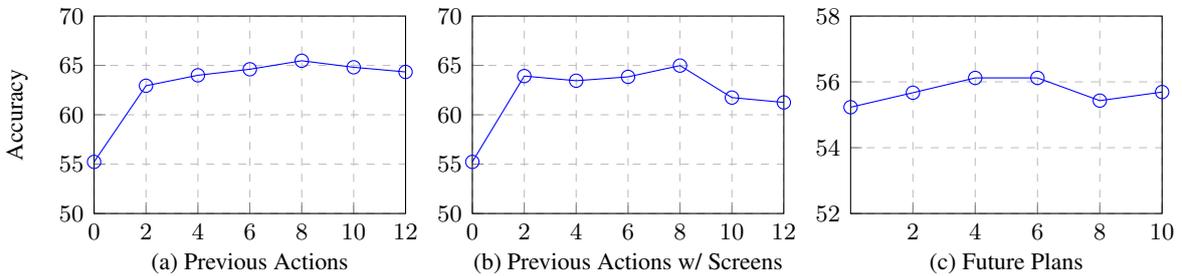
\begin{figure*}[htb]
\centering
    \begin{tikzpicture}
        \pgfplotsset{small, samples=10}
        \begin{groupplot}[
            group style = {group size = 3 by 1, horizontal sep = 25pt}, 
            width = 5.68cm, 
            height = 4.2cm]
            \nextgroupplot[
                align = center,
                legend style={at={(0.5,1.5)},anchor=north},
                xmin=0, xmax=12,
                ymin=50, ymax=70,
                xtick={0, 2, 4, 6, 8, 10, 12},
                xlabel={(a) Previous Actions},
                ylabel style={align=center},
                ylabel={Accuracy},
                ytick={50, 55, 60, 65, 70},
                ymajorgrids=true,
                xmajorgrids=true,
                grid style=dashed,
                x label style={at={(axis description cs:0.5,-0.15)},anchor=north},
                y label style={at={(axis description cs:-0.18,0.5)},anchor=south},
                xtick pos=bottom,
                ytick pos=left,
                ]
                \addplot[
                    color=blue,
                    mark=o,
                    mark size=2.5pt,
                    ]
                    coordinates {
                    (0, 55.23)
                    (2, 62.94)
                    (4, 64.00)
                    (6, 64.61)
                    (8, 65.47)
                    (10, 64.81)
                    (12, 64.34)
                    };
            \nextgroupplot[
                align = center,
                legend style={at={(0.5,1.5)},anchor=north},
                xmin=0, xmax=12,
                ymin=50, ymax=70,
                xtick={0, 2, 4, 6, 8, 10, 12},
                xlabel={(b) Previous Actions w/ Screens},
                ylabel style={align=center},
                ytick={50, 55, 60, 65, 70},
                ymajorgrids=true,
                xmajorgrids=true,
                grid style=dashed,
                x label style={at={(axis description cs:0.5,-0.15)},anchor=north},
                y label style={at={(axis description cs:-0.18,0.5)},anchor=south},
                xtick pos=bottom,
                ytick pos=left,
                ]
                \addplot[
                    color=blue,
                    mark=o,
                    mark size=2.5pt,
                    ]
                    coordinates {
                    (0, 55.23)
                    (2, 63.92)
                    (4, 63.45)
                    (6, 63.84)
                    (8, 64.98)
                    (10, 61.74)
                    (12, 61.25)
                    };
            \nextgroupplot[
                align = center,
                legend style={at={(0.5,1.28)},anchor=south},
                xmin=0, xmax=10,
                ymin=52, ymax=58,
                xtick={2, 4, 6, 8, 10},
                xlabel={(c) Future Plans},
                ylabel style={align=center},
                ytick={52, 54, 56, 58},
                ymajorgrids=true,
                xmajorgrids=true,
                grid style=dashed,
                x label style={at={(axis description cs:0.5,-0.15)},anchor=north},
                y label style={at={(axis description cs:-0.18,0.5)},anchor=south},
                xtick pos=bottom,
                ytick pos=left,
                legend columns=2 row=1,
                ]
                \addplot[
                    color=blue,
                    mark=o,
                    mark size=2.5pt,
                    ]	
                    coordinates {
                    (0, 55.23)
                    (2, 55.67)
                    (4, 56.12)
                    (6, 56.12)
                    (8, 55.43)
                    (10, 55.69)
                    };
        \end{groupplot}
    \end{tikzpicture}
    \vspace{-3mm}
    \caption{Performance of Auto-GUI with respect to varying length of chain of actions.}
    \vspace{-1mm}
    \label{fig:n-shot-ablation}
    
\end{figure*}

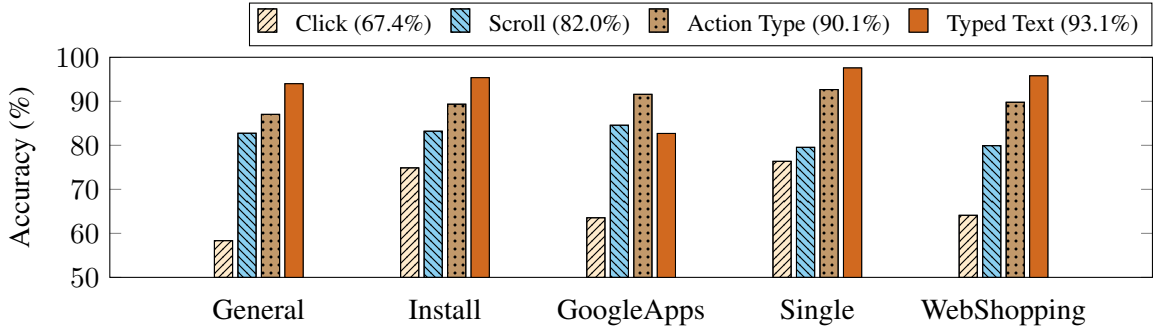
\begin{figure*}[htb]
  {\centering
  \pgfplotsset{compat=1.13,
    /pgfplots/ybar legend/.style={
    /pgfplots/legend image code/.code={%
       \draw[##1,/tikz/.cd,yshift=-0.25em]
        (0cm,0cm) rectangle (7pt,0.8em);},
   },
}
  \pgfplotsset{width=15.3cm, height=4.5cm, compat=1.3}
    \begin{tikzpicture}  
        \begin{axis}  
        [  
            ybar,
            ymin=50, ymax=100,
            ytick={50,60,70,80,90,100},
            major x tick style = transparent,
            bar width=6.8pt,
            legend columns=4 row=1,
            enlarge x limits=0.2,
            ylabel={Accuracy (\%)},
            symbolic x coords={2,4,6,8,10},  
            xticklabels={General, Install, GoogleApps, Single, WebShopping},
            xtick=data,  
        legend cell align=left,
legend style={
                        at={(1,1.05)},
                        anchor=south east,
                        column sep=1ex,
                        font=\footnotesize,
                },
            ]  
	
        \addplot[ybar, fill=blanchedalmond,  postaction={pattern=north east lines}] coordinates {
            (2, 58.34)(4, 74.88)(6, 63.54) (8, 76.36)(10, 64.1)
        };  
        \addplot[ybar, fill=babyblue,  postaction={pattern=north west lines}] coordinates {
          (2, 82.74)(4, 83.2)(6, 84.56) (8, 79.54)(10, 79.92)
        };  
        \addplot[ybar, fill=camel,  postaction={pattern=dots}]  coordinates {
            (2, 87.03)(4, 89.34)(6, 91.58) (8, 92.67)(10, 89.78)
        };  
        \addplot[ybar, fill=cinnamon,  postaction={}] coordinates {
          (2, 93.99)(4, 95.36)(6, 82.69) (8, 97.57)(10, 95.78)
        };  
        \legend{\small{Click (67.4\%)}, \small{Scroll (82.0\%)},\small{Action Type (90.1\%)},\small{Typed Text (93.1\%)}
        }
        \end{axis}  
    \end{tikzpicture}  
    \vspace{-2mm}
   
}
  \caption{{Category accuracy of Auto-GUI. Values in parentheses represent the average accuracy on the subsets. }}\label{tab:one-demo}
  \vspace{-3mm}
\end{figure*}

\begin{table}[t]
\centering\small
\vspace{2mm}
\renewcommand\tabcolsep{6.8pt} 
\begin{tabular}{lc} 
\toprule
 Model & Accuracy \\
\midrule
Auto-GUI & \textbf{74.27} \\
\midrule
\quad w/o future action plan  & 73.78 \\
\quad w/o previous action history & 68.81 \\
\quad w/o chain of actions & 68.53 \\
\midrule
\quad w/o coordinate normalization & 70.23 \\
 \bottomrule
\end{tabular}
\caption{Ablation study of Auto-GUI.}
 \label{tab:ablation_results}
 \vspace{-3mm}
\end{table}

\subsection{Main Results}
Table \ref{tab:main_results} shows the main results. Based on the results, we have the following observations.

(i) Auto-GUI$_\text{unified}$ achieves the best overall performance compared with all the baselines. Compared with separate (not unified) models, Auto-GUI$_\text{unified}$ shows general effectiveness across various tasks. The results show that a unified multimodal model out of \emph{first principles thinking} can serve as a strong autonomous agent. Compared with previous BC models, Auto-GUI$_\text{unified}$ has two major advantages. First, Auto-GUI$_\text{unified}$ is a unified model that can be adapted to different scenarios without the need to train specific models for each task. Second, Auto-GUI$_\text{unified}$ does not need additional annotations and is more practical in real-world applications. Furthermore, Auto-GUI yields divergent performance across subsets. We provide the explanation in Appendix \ref{appendix:subset} to save space.

(ii) Both the chain of actions and coordinate normalization contribute to the overall performance (+5.74\% and 4.04\%, respectively), as evidenced by the ablation study in Table \ref{tab:ablation_results}. Additionally, we set the maximum numbers of the previous actions and future actions to 8 and 4, respectively. The choice is made according to our analysis of the General subset with Auto-GUI$_\text{separate}$ (Figure \ref{fig:n-shot-ablation}). The model under those setups achieves the optimal performance, and neither the input nor output sequence lengths exceed the model limit.

(iii) For the LLMs, using either prompting or fine-tuning techniques does not achieve competitive performance compared with multimodal approaches. The most plausible reason is that they learn from the parsed HTML elements of the screen so that they may suffer from information loss compared with more informative vision features of the screens. {Specifically, we find that ChatGPT is quite accurate at predicting the action type but fails at lower-level executions (Appendix \ref{appendix:cat_icl})}. 


\section{Analysis}
\subsection{Analysis of Auto-GUI by Category}\label{sec:category}
To dive into the capability of Auto-GUI, we calculate the click region accuracy, scroll direction accuracy, action type accuracy, and typed text accuracy. Figure \ref{tab:one-demo} presents the results. We see that Auto-GUI achieves over 90\% action type accuracy on average. In contrast, the major challenges lie within the click region and scroll direction predictions. Although the model is able to predict the right action most of the time, it tends to click a wrong place or scroll in a wrong direction. The result reveals a future direction of improving the model's ability to understand the screen layouts, e.g., using more advanced vision features. 

\begin{table*}[htb]
\centering\small
\vspace{-3mm}
\renewcommand\tabcolsep{6pt} 
\begin{tabular}{l|c|ccccc} 
\toprule
 Model & Overall & General & Install & GoogleApps & Single & WebShopping \\
\midrule
Auto-GUI on CLIP 	& 71.84	& 66.28	& 	74.40		& 69.71		& 81.60		& 67.23 \\
Auto-GUI on BLIP-2 	&  {74.27}	& {68.24} &	{76.89} & {71.37} & {84.58} & {70.26} \\
\midrule
Auto-GUI on Vanilla-T5$_\text{large}$ & 72.98	& 66.61	& 75.40	& 70.86	& 83.47	& 68.54\\
Auto-GUI on FLAN-T5$_\text{large}$ 	&  73.36		& 67.59		& 76.35		& 70.71	& 	83.01		& 69.12 \\
Auto-GUI on FLAN-Alpaca$_\text{large}$ 	&  {74.27}	& {68.24} &	{76.89} & {71.37} & {84.58} & {70.26} \\
\midrule
Auto-GUI on FLAN-Alpaca$_\text{small}$ 	& 71.38	& 	65.26	& 	74.90		& 68.70	& 	81.20	& 	66.83 \\
Auto-GUI on FLAN-Alpaca$_\text{base}$	&  72.84	& 	66.97	& 	75.93	& 	70.29	& 	82.56	& 	68.46 \\
Auto-GUI on FLAN-Alpaca$_\text{large}$ 	&  {74.27}	& {68.24} &	{76.89} & {71.37} & {84.58} & {70.26} \\
 \bottomrule
\end{tabular}
\caption{Results varying vision features and pre-trained language model weights.}
 \label{tab:category_results}
\end{table*}

\begin{table*}[htb]
\centering\small
\vspace{-1mm}
\renewcommand\tabcolsep{12pt} 
\begin{tabular}{l|c|c|c} 
\toprule
 Model & Feature Extraction (s/n) & Model Inference (s/n) & Peak GPU Memory (GB)\\
\midrule
Auto-GUI$_\text{base}$  & 0.06 & 0.19 (45x) & 4.6 (10x) \\
Auto-GUI$_\text{large}$ & 0.06 & 0.59 (15x) & 8.2 (6x) \\
\midrule
Llama 2 	& - & 8.5 &  49.7 \\
 \bottomrule
\end{tabular}
\caption{Computation cost of Auto-GUI and Llama. ``s/n'' is computed by time (s) divided by the number of inferences (n). Llama 2 is hosted with 8-bit quantization and float16 precision to improve the inference speed.}
 \label{tab:efficiency}
 \vspace{-3mm}
\end{table*}

\begin{figure}[t]
   \begin{center}
   \vspace{2mm}
  \includegraphics[width=0.5\textwidth]{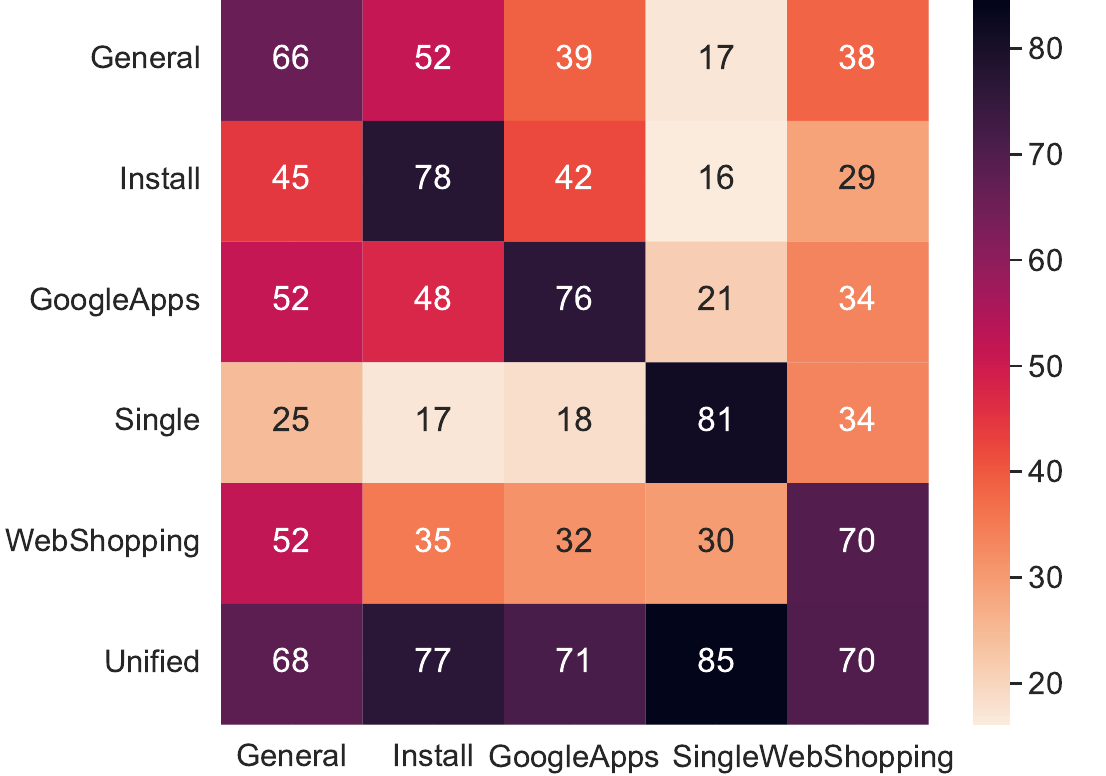}
  \end{center}
  \vspace{-3mm}
  \caption{Transfer results of Auto-GUI. The numbers in the cells mean test accuracy. For example, we train an Auto-GUI model on the training set of General and then test its performance on the tests of each subset.}
  \label{fig:heatmap}
  \vspace{-4.5mm}
\end{figure}

\subsection{Comparison with ICL by Category}\label{appendix:cat_icl}
{To understand how the ICL baseline performs on our task and assess the advantage of Auto-GUI, we conduct a category comparison with ChatGPT.} 

\begin{table}[htb]
\centering\small
\renewcommand\tabcolsep{6pt} 
\begin{tabular}{l|c|ccccc} 
\toprule
 Model & Overall & Action Type & Click & Scroll  \\
\midrule
ChatGPT & 5.93  & 41.72   & 8.50 & 4.00  \\
Auto-GUI & 68.24  & 87.03   & 58.34 & 82.74  \\
\bottomrule
\end{tabular}
\caption{Category comparison with the ICL baseline on the General test set.}
 \label{tab:app_icl_compare}
\end{table}

{In Table \ref{tab:app_icl_compare}, we see that the ICL method (ChatGPT) is quite accurate at predicting the action type (41.72\%) but fails at lower-level executions, e.g., clicking positions (8.5\%) and scrolling directions (4.0\%). The results show that using HTML-based layout information is not enough to accurately execute actions. In contrast, Auto-GUI has the advantage of predicting both action types and performing low-level executions by leveraging multimodal perception and the chain-of-action technique.}

\subsection{Generalization Ability}\label{sec:general}
As our approach is designed under first principles thinking and does not rely on pre-defined internal APIs, it could be easily generalized to new task domains. To verify the generality, we evaluate the performance of Auto-GUI$_\text{separate}$ on each subset in Figure \ref{fig:heatmap}. For example, we train an Auto-GUI$_\text{separate}$ model on the training set of General and then test its performance on the tests of each subset. 

We see that our approach is able to achieve a decent performance, though the domains vary. This result reveals that
the model could capture general knowledge for the GUI control task; thus is applicable to different domains. In addition, the unified model Auto-GUI$_\text{unified}$ can serve as a potential choice in real-world applications owing to more coverage of training data.

\subsection{Comprehensive Analysis}
Here, we present a comprehensive analysis of the choice of pre-trained features and model scale. The results are summarized in Table \ref{tab:category_results}.

(i) Pre-trained Features. There are two kinds of pre-trained features used in this work, the vision features and language model weights. For vision features, we compare two popular types, CLIP \citep{radford2021learning} and BLIP-2 \citep{li2023blip}. We observe that BLIP-2 achieves {relatively better} performance. Therefore, we use BLIP-2 by default in Auto-GUI. For pre-trained language model weights, we compare initializing the model with the vanilla T5 \citep{raffel2020exploring}, FLAN-T5 \citep{chung2022scaling}, and FLAN-Alpaca \citep{taori2023alpaca} weights under the large size. We see that FLAN-Alpaca achieves the best performance as it has been optimized with Stanford Alpaca synthetic instruction tuning data.

(ii) Model Scale. Compared with the performance gains from our technique components (chain of actions and coordinate normalization) in Table \ref{tab:ablation_results}, the benefit of scaling parameter size becomes relatively marginal. As we observe that a larger model size does not lead to dramatic improvement in performance, we do not scale the model scale but focus on the base (220M) and large (770M) models in this work. In addition, our choice is also based on other considerations, including the constriction of GPU memory and computation budget. 

According to the analysis above, our work offers insights into the determinants of model efficacy:

$\bullet$ Vision features are critical, underscoring the significance of effective perception.

$\bullet$ Model scale exhibits diminished importance, indicating that larger model sizes do not necessarily yield dramatic performance enhancements.

\subsection{Computation Cost}\label{sec:efficiency}

Table \ref{tab:efficiency} compares the inference speed and GPU memory cost for Auto-GUI and Llama 2. Auto-GUI is able to achieve nearly real-time inference (within less than one second for an action prediction) with less than 10GB GPU memory. The inference speed is over 10 times faster than Llama 2. Our work shows the strength of the medium-sized language model in building autonomous agents, which is able to achieve competitive performance with fast inference speed and modest resource cost.

\section{Conclusion}
This work presents an autonomous GUI agent called Auto-GUI that can interact in a multimodal GUI environment without environment parsing or application-dependent API access. In addition, we propose a chain-of-action technique that leverages the previously executed actions and future action plans to help the agent decide what action to execute. 
Experimental results show that Auto-GUI achieves superior performance to previous prompting-based and fine-tuning baselines. We show that it is possible to achieve state-of-the-art performance by an end-to-end model without relying on external tools and application-specific APIs to parse the environment and interpret the predicted actions.
Besides the strong performance and generality across domains, Auto-GUI is able to infer an action in less than one second.

\section*{Limitations}
We acknowledge two primary limitations in our study.  First, we opted not to extend the approach to extremely large models because our work aims to provide a simple yet effective solution for GUI agents. Our findings suggest that scaling may not be fundamentally advantageous in GUI problems. The significance of model scale tends to diminish---increasing the model size does not necessarily result in a substantial performance enhancement. Second, our experiments and analysis were exclusively conducted on AITW, which is the largest-scale and widely recognized benchmark dataset in the research line of autonomous GUI agents, to provide timely and pertinent insights. Given the rapid development of the field, we anticipate future studies to explore the application of our approach on other benchmark datasets as they become available. 

\bibliography{custom}

\clearpage
\appendix
\section{Data Details}

\subsection{Data Examples}\label{app:example}
\begin{figure}[htb]
  \begin{center}
 \includegraphics[width=1.0\columnwidth]{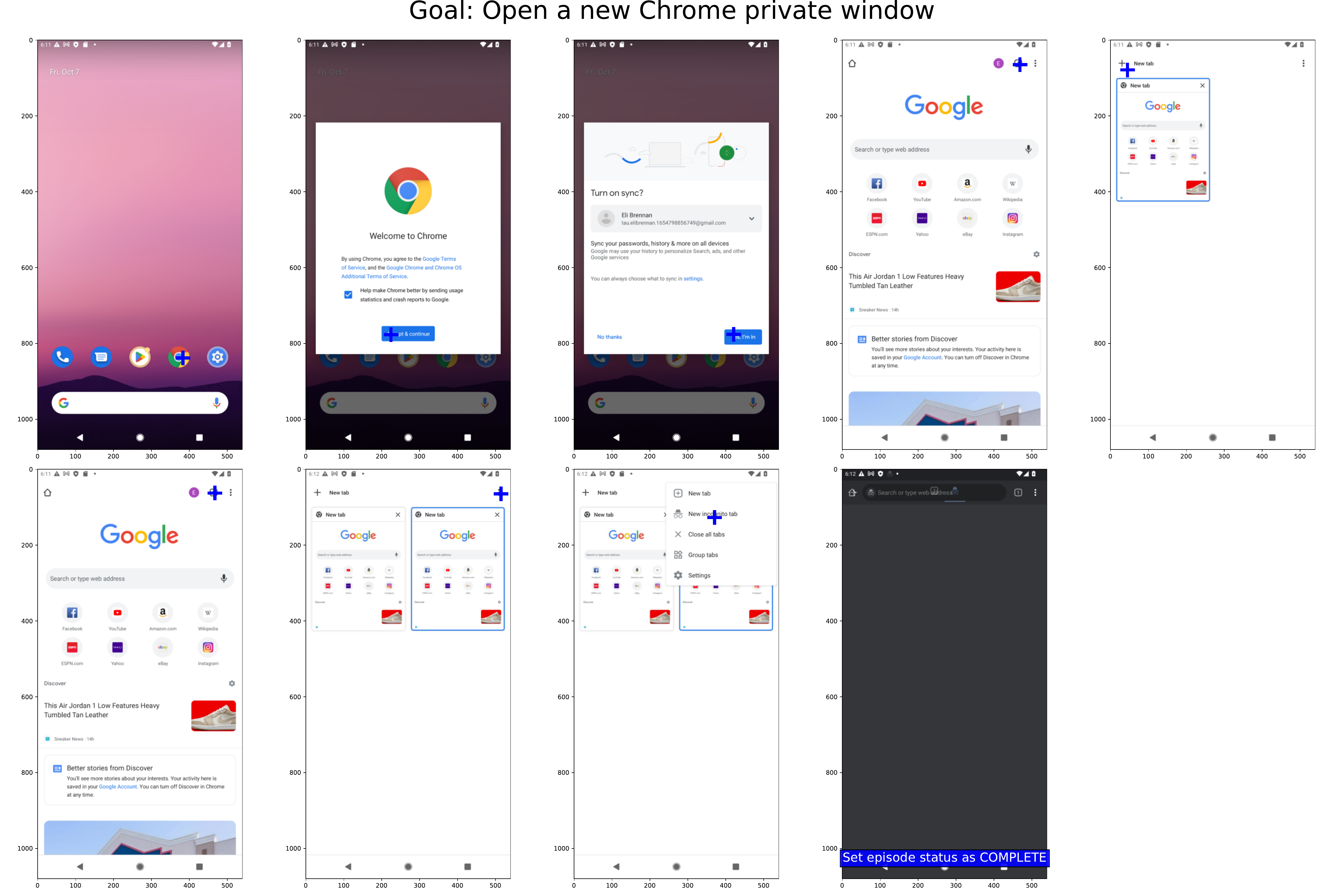}
  \end{center}
  \caption{An example episode from General.}
  \label{fig:data_general}
\end{figure}

\begin{figure}[htb]
  \begin{center}
 \includegraphics[width=1.0\columnwidth]{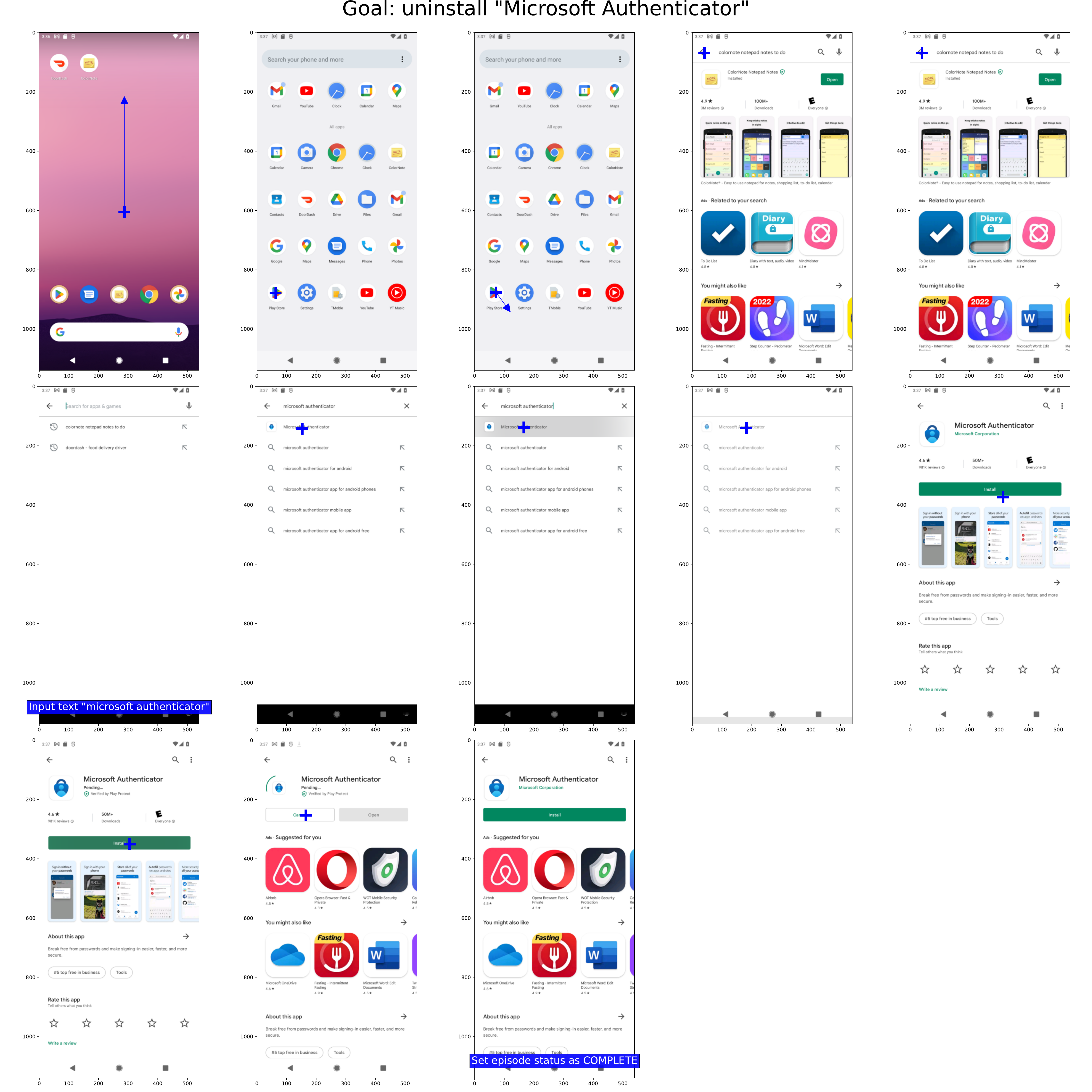}
  \end{center}
  \caption{An example episode from Install.}
  \label{fig:data_install}
\end{figure}

\begin{figure}[htb]
  \begin{center}
 \includegraphics[width=1.0\columnwidth]{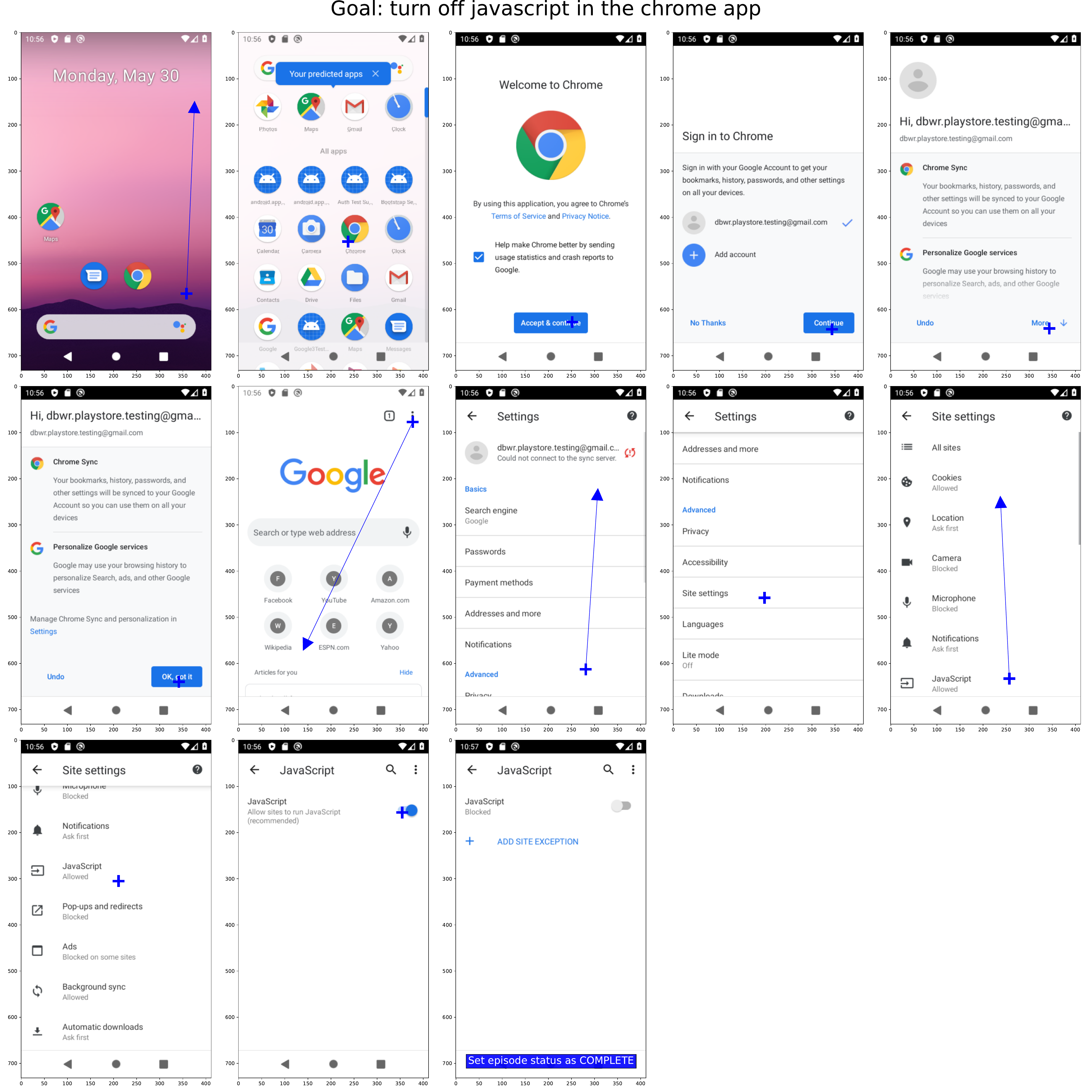}
  \end{center}
  \caption{An example episode from GoogleApps.}
  \label{fig:data_google}
\end{figure}

\begin{figure}[htb]
  \begin{center}
 \includegraphics[width=1.0\columnwidth]{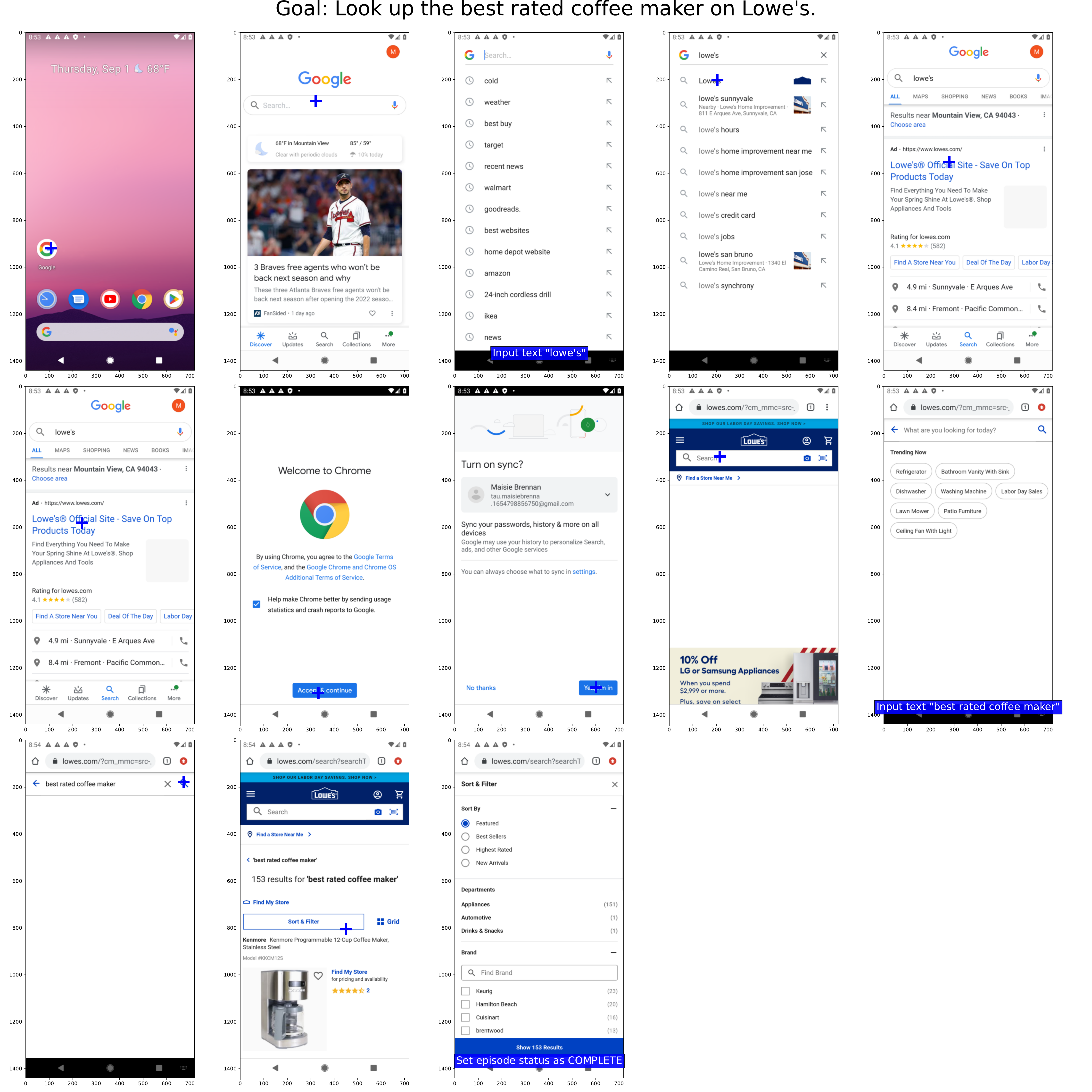}
  \end{center}
  \caption{An example episode from WebShopping.}
  \label{fig:data_web}
\end{figure}

We show the data examples from the AITW benchmark dataset \citep{rawles2023android}.
Figures \ref{fig:data_general}-\ref{fig:data_web} show the examples in each subset, i.e., General, Install, GoogleApps, Single, and WebShopping. The gold actions for each screen are depicted in the illustrations for reference.

\subsection{Data Statistics}\label{appendix:data}
We use the AITW benchmark dataset \citep{rawles2023android}. AITW is a large-scale benchmark dataset for GUI control, which contains natural language instructions, screenshots, and actions. There are 715$K$ episodes spanning 30$K$ unique instructions, covering diverse multi-step tasks such as application operation, web searching, and web shopping, on over 350 Apps and websites. This dataset covers various device types and operation systems in varying screen resolutions to ensure generality. There are five subsets in the benchmark dataset: General, Install, GoogleApps, Single, and WebShopping. 

(i) General contains miscellaneous tasks that need interaction with third-party Apps and websites, as well as question answering.

(ii) Install contains tasks related to installing, uninstalling, logging Apps, and App login support.

(iii) GoogleApps contains tasks about manipulating various Google applications such as Gmail, Calendar, Photos, and Settings.

(iv) Single contains atomic tasks (e.g., ``upvote the post'') whose preceding actions have been already completed (e.g., opening Instagram, going to home feed, looking at a post). 

(v) WebShopping contains tasks related to online shopping on E-commerce websites, e.g., searching for an item, adding an item to the cart, and viewing the shopping cart.

Table \ref{tab:data} presents the data statistics of the AITW dataset. Each subset is split episode-wise into a training, validation, and test set (80/10/10\%). 

\begin{table}[!htb]
    \centering\small
    \begin{tabular}{lllc}
    \toprule
     {Dataset}  & Episodes & Screens  &  Instructions\\
    \midrule
     General & 9,476 & 85,413 & 545 \\
     Install &  25,760 & 250,058 & 688 \\
     GoogleApps &  625,542 & 4,903,601 & 306  \\
     Single & 26,303 & 85,668 & 15,366 \\
     WebShopping & 28,061 & 365,253 & 13,473 \\
    \bottomrule
  \end{tabular}
  \caption{Dataset statistics\label{tab:data}.}
\end{table}

\begin{table*}[t]
\centering\small
\renewcommand\tabcolsep{6pt} 
\begin{tabular}{p{3.6cm}p{10cm}} 
\toprule
 Action Type & Target Output \\
\midrule
dual-point gesture (click) 	& ``action\_type'': 4, ``touch\_point'': [0.8497, 0.5964], ``lift\_point'': [0.8497, 0.5964], ``typed\_text'': ``'' \\
dual-point gesture (scroll) 	& ``action\_type'': 4, ``touch\_point'': [0.2, 0.5], ``lift\_point'': [0.8, 0.5], ``typed\_text'': ``'' \\
type 	& ``action\_type'': 3, ``touch\_point'': [-1.0, -1.0], ``lift\_point'': [-1.0, -1.0], ``typed\_text'': ``what's the news in chile?'' \\
go\_back 	& ``action\_type'': 5, ``touch\_point'': [-1.0, -1.0], ``lift\_point'': [-1.0, -1.0], ``typed\_text'': ``'' \\
go\_home 	& ``action\_type'': 6, ``touch\_point'': [-1.0, -1.0], ``lift\_point'': [-1.0, -1.0], ``typed\_text'': ``'' \\
enter 	& ``action\_type'': 7, ``touch\_point'': [-1.0, -1.0], ``lift\_point'': [-1.0, -1.0], ``typed\_text'': ``'' \\
status\_complete 	& ``action\_type'': 10, ``touch\_point'': [-1.0, -1.0], ``lift\_point'': [-1.0, -1.0], ``typed\_text'': ``'' \\
 \bottomrule
\end{tabular}
\caption{Target output examples after the coordinate normalization.}
 \label{tab:norm_exp}
\end{table*}

\section{Implementation Details}
\subsection{Coordinate Normalization}\label{app:norm}
Recall that a target action consists of four components: action type, touch point, lift point, and typed text. We consider six action types: \textit{dual-point gesture}, \textit{type}, \textit{go\_back}, \textit{go\_home}, \textit{enter}, and \textit{status\_complete}.  A dual-point gesture comprises a touch point and a lift point with $[y, x]$ coordinates. The gesture actions ensure a flexible action space and can represent clicks and scrolls at arbitrary locations. 
For example, a gesture action \{``touch\_point'': [0.7761, 0.7089], ``lift\_point'': [0.7761, 0.7089]\} means clicking at the coordinate [0.7761, 0.7089], while a gesture action \{``touch\_point'': [0.1898, 0.4477], ``lift\_point'': [0.8242, 0.4077]\} means scrolling down. A type action means typing a text and the text is placed in the <typed\_text> field. The other action types, i.e., go\_back, go\_home, enter, and status\_complete are system actions, whose corresponding <touch\_point>, <lift\_point> fields are filled with -1, and the <typed\_text> is empty.

We observe that high-precision coordinates are not necessary for representing a click or scroll action. Therefore, we apply normalized values of the coordinates, which helps accelerate convergence and mitigate the ambiguity of coordinates. The normalization is applied to click and scroll actions. For click actions, we keep four decimal places. For scroll actions, we first determine the scroll direction with the touch and lift points. Then, we transform the touch and lift points into fixed directional coordinates as follows: ``up'': \{[0.8, 0.5], [0.2, 0.5]\},
``down'': \{[0.2, 0.5], [0.8, 0.5]\},
``left'': \{[0.5, 0.8], [0.5, 0.2]\},
``right'': \{[0.5, 0.2], [0.5, 0.8]\}, where \{[$\cdot$], [$\cdot$]\} consists of the touch point and lift point in the first [$\cdot$] and second [$\cdot$]. We provide examples of target actions in Table \ref{tab:norm_exp}.

\begin{table*}[htb]
\centering\small
\vspace{-2mm}
\renewcommand\tabcolsep{8.8pt} 
\begin{tabular}{l|c|ccccc} 
\toprule
 Model & Overall & General & Install & GoogleApps & Single & WebShopping \\
\midrule
Auto-GUI & \textbf{74.27}	&\textbf{68.24}	&\textbf{76.89}	&\textbf{71.37}	&\textbf{84.58}	&\textbf{70.26} \\
\midrule
\quad w/o  previous action history & 73.78 &	67.97	&76.66	&71.00	&83.64	&69.62 \\
\quad w/o  future action plan & 68.81 	&59.01	&72.34	&67.95	&81.53	&63.24 \\
\quad w/o chain of actions & 68.53	&58.99	&72.06	&67.50	&81.25	&62.86 \\
\midrule
\quad w/o coordinate normalization & 70.23	&63.79	&73.28	&66.63	&82.11	&65.33 \\
 \bottomrule
\end{tabular}
\caption{Ablation study of Auto-GUI design components. We adopt Auto-GUI$_\text{unified}$ for analysis.}
 \label{tab:ablation_results_more}
 \vspace{-2mm}
\end{table*}

\begin{table*}[htb]
\centering\small
\renewcommand\tabcolsep{10.2pt} 
\begin{tabular}{l|c|ccccc} 
\toprule
 Model & Overall & General & Install & GoogleApps & Single & WebShopping \\
\midrule
Auto-GUI$_\text{base}$ 	&  72.84	& 	66.97	& 	75.93	& 	70.29	& 	82.56	& 	68.46 \\
\quad w/ Screen Descriptions & 75.54 & 70.30 & 78.05	& 73.04 & 85.31 & 71.00\\
 \bottomrule
\end{tabular}
\caption{Results of Auto-GUI when using annotated screen descriptions.}
 \label{tab:anno_results}
\end{table*}

\subsection{Baselines}\label{app:baselines}
We adopt three types of baselines for comparisons. {The baselines encompass the in-context earning (ICL) and fine-tuning paradigms, along with various backbone models of different sizes. This choice of baselines allows for a comprehensive comparison with our proposed approach.}

{(i) In-context Learning LLMs. Few-shot PaLM 2, ChatGPT (turbo-3.5) are adopted. Following previous studies \citep{rawles2023android,wang2023enabling}, we feed the LLM a textual description of the screen and a user instruction.  The textual description of the screen is formatted as an HTML syntax, providing the information of GUI elements derived from OCR detection and icon detection from external tools \citep{rawles2023android}. The model is required to predict an action among pre-defined actions. If the action is clicking, the model will be required to provide the index of the clicked GUI element. Alternatively, the model needs to provide the scroll direction if the action is scrolling. In addition, 5-shot CoT prompting is leveraged to improve the performance (Appendix \ref{app:prompt}). In addition, we report the results of the multimodal GPT-4V by taking the vision image and action history as the input based on \citet{yan2023gpt}.}

{(ii) Fine-tuned LLMs. We adopt Llama 2 \citep{touvron2023llama} as the baseline and fine-tune it with LoRA. We feed the model with the user instruction and the screen descriptions in HTML syntax (the same as adopted for in-context learning LLMs). The model is expected to predict the action in the same output format as in-context learning LLMs. As fine-tuning an LLM is expensive, we randomly sample 1\% training data to help the LLM adapt to our tasks.}

{(iii) Specialized GUI Agent. We adopted the Behavioural Cloning (BC) agent, which reported the state-of-the-art performance in \citet{rawles2023android}. BC is a Transformer-based architecture that takes a task instruction, the current screen, and a stacked history of screen observations and actions as input. The task instruction and OCR-detected texts are encoded by a pre-trained BERT. The icons are represented by the embeddings for each of the bounding box points. The screen history is modeled by the $\{x,y\}$ positions of the touch and lift actions. All the embedded representations are fused to predict the action by a decoder. There are two BC variants, BC-single and BC-history, depending on whether the model takes as input the screen-action history.}

\subsection{LLM Prompt}\label{app:prompt}
\begin{figure*}[h!]
\centering
\includegraphics[width=\textwidth]{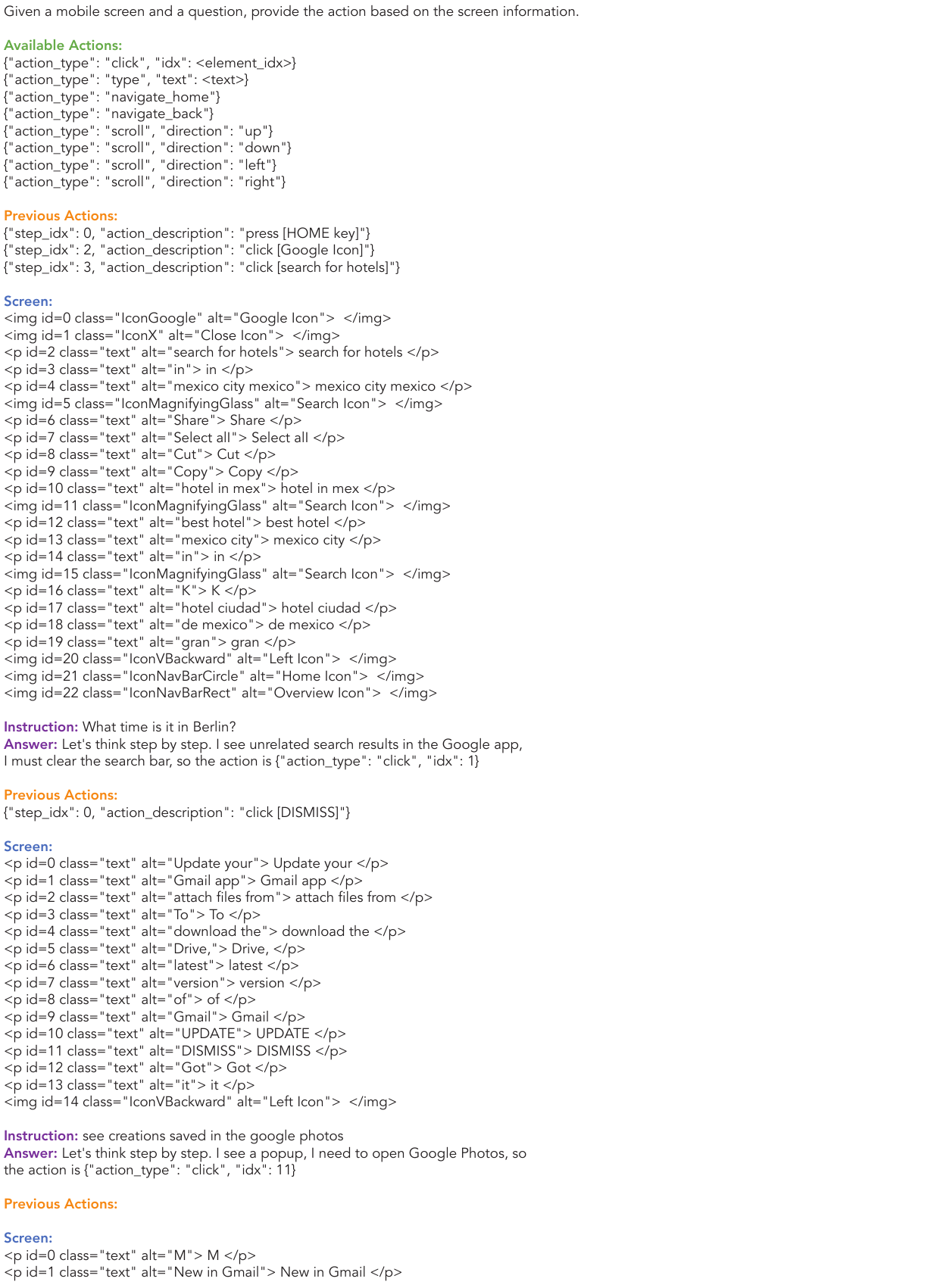}
\caption{LLM Prompt (Part-I).}
\label{fig:prompt-1}
\end{figure*}

\begin{figure*}[h!]
\centering
\includegraphics[width=\textwidth]{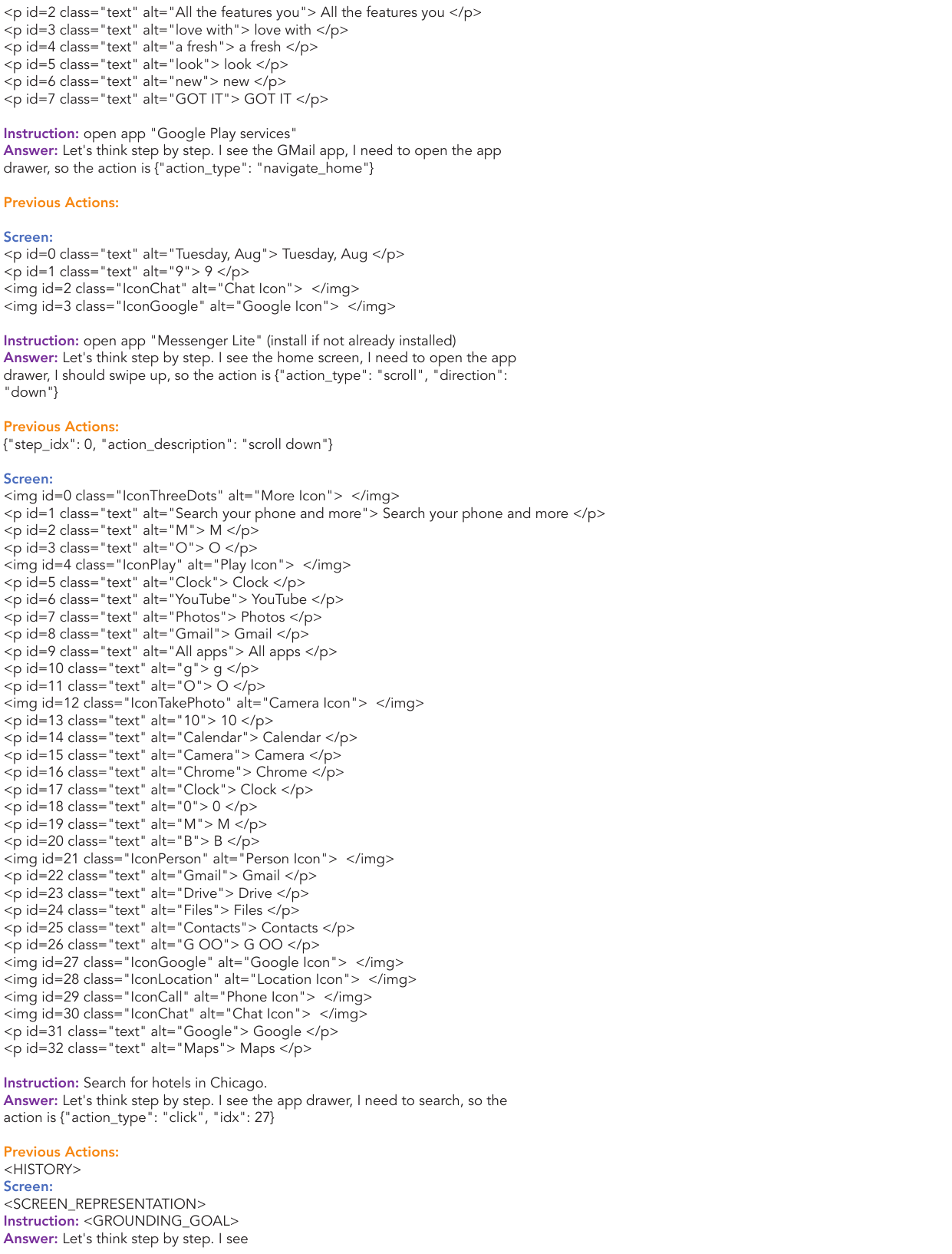}
\caption{LLM Prompt (Part-II).}
\label{fig:prompt-2}
\end{figure*}

We use the prompt in Figures \ref{fig:prompt-1}-\ref{fig:prompt-2} for PaLM 2-CoT and ChatGPT-CoT owing to its optimal performance reported in \citet{rawles2023android}.

\section{Further Analysis}
\subsection{Subset Analysis}\label{appendix:subset}
We notice that Auto-GUI$_\text{unified}$ performs relatively inferior to BC-history on the two App-centered subsets, Install and GoogleApps. It is reasonable because we only use 10\% training data of GoogleApps considering the data balance and computation overhead. We observe that the performance does not improve when we use all the training data of GoogleApps, possibly due to the data imbalance issue \citep{zhang2022task}. In contrast, our separate model Auto-GUI$_\text{separate}$ can achieve better performance than BC-history, showing that our approach is better than BC-history under the same training setting. As we aim to study a simple and unified approach that achieves generally strong performance, we leave the treatment of the data imbalance issue in future work.

\subsection{Ablation Study}\label{appendix:abl}

Table \ref{tab:ablation_results_more} shows the detailed results of the ablation study. We see that both the chain of actions and coordinate normalization contribute to the overall performance (+5.74\% and 4.04\%, respectively).

\subsection{Using Screen Descriptions}
We are interested in whether Auto-GUI can be further improved when screen annotations are available. Therefore, we incorporate screen descriptions containing icon and text information, organized in HTML syntax, into our language input $X_{\textrm{language}}$. Detailed examples of screen descriptions can be found in the ``Screen'' block in Appendix \ref{app:prompt}.

In Table \ref{tab:anno_results}, we see that Auto-GUI can perform better when the annotated screen descriptions are available. The results show that there is still room for performance gains for Auto-GUI. However, as the annotations are not always available in real-world applications, we do not include them by default in our framework.


\end{document}